\documentclass[final]{cvpr}

\usepackage{times}
\usepackage{epsfig}
\usepackage{graphicx}
\usepackage{amsmath}
\usepackage{amssymb}

\usepackage[pagebackref=true,breaklinks=true,colorlinks,bookmarks=false]{hyperref}

\def\cvprPaperID{5085} 
\def\confYear{CVPR 2021}

\begin{document}

\title{Spatial-Temporal Correlation and Topology Learning \\ for Person Re-Identification in Videos}


\author{Jiawei Liu, Zheng-Jun Zha\textsuperscript{*}, Wei Wu, Kecheng Zheng, Qibin Sun\\
	University of Science and Technology of China, China\\
	{\tt\small \{jwliu6,zhazj,qibinsun\}@ustc.edu.cn, \{wuvy,zkcys001\}@mail.ustc.edu.cn}
}
\maketitle

\pagestyle{empty}  
\thispagestyle{empty} 
\begin{abstract}
Video-based person re-identification aims to match pedestrians from video sequences across non-overlapping camera views. The key factor for video person re-identification is to effectively exploit both spatial and temporal clues from video sequences. In this work, we propose a novel Spatial-Temporal Correlation and Topology Learning framework (CTL) to pursue discriminative and robust representation by modeling cross-scale spatial-temporal correlation. Specifically, CTL utilizes a CNN backbone and a key-points estimator to extract semantic local features from human body at multiple granularities as graph nodes. It explores a context-reinforced topology to construct multi-scale graphs by considering both global contextual information and physical connections of human body. Moreover, a 3D graph convolution and a cross-scale graph convolution are designed, which facilitate direct cross-spacetime and cross-scale information propagation for capturing hierarchical spatial-temporal dependencies and structural information. By jointly performing the two convolutions, CTL effectively mines comprehensive clues that are complementary with appearance information to enhance representational capacity. Extensive experiments on two  video benchmarks have demonstrated the effectiveness of the proposed method and the state-of-the-art performance.
\end{abstract}

\section{Introduction}

\footnotetext{* Corresponding author.}
Person re-identification (Re-ID) is an important technology to retrieve a person-of-interest across non-overlapping cameras. It has drawn increasing attention during the past few years, owing to its broad application in many realistic scenarios, such as video surveillance \cite{gu2020appearance,zha2020adversarial} and behavior analysis \cite{wu2020adaptive} \textit{etc}. However, this task remains challenging due to the variations in illumination, viewpoint and pose, as well as the influence of background clutter and occlusion.

\begin{figure}[!t]
	\begin{center}
		\includegraphics[width=0.80 \linewidth]{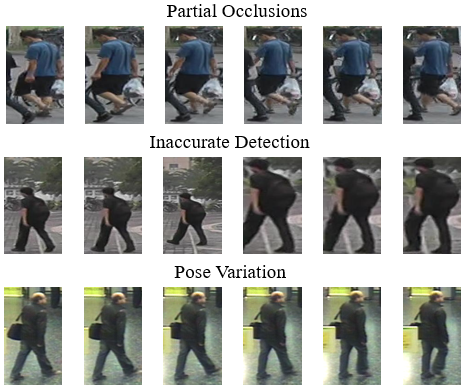}
	\end{center}
	\caption{Three example video sequences on MARS and iLIDS-VID datasets with partial occlusions, inaccurate detection and viewpoint variation.}
	\label{img1}
\end{figure}
Existing person Re-ID approaches are mainly divided into two categories: image-based methods \cite{sun2018beyond,shen2018person,tay2019aanet,liu2019adaptive} and video-based methods \cite{Subramaniam_2019_ICCV,liu2019dense,li2020relation}. The former exploits static images without temporal information to retrieve pedestrians. It has achieved impressive advances with the surge of deep learning technique in recent years \cite{jiang2019rethinking}. However, image-based person Re-ID heavily relies on the quality of static images, which are sensitive to noise, occlusion and viewpoint variation, \textit{etc}. Different from static images with limited content, video sequences contain rich spatial-temporal information across a long span of time, which can provide clean and informative clues against these problem \cite{li2020relation,fu2019sta}. Thus, video-based person Re-ID has the potential to solve the restrictions in image-based person Re-ID.


A typical video-based person Re-ID pipeline extracts and aggregates spatial and temporal clues from video sequences to generate discriminative representations. Some preliminary methods \cite{mclaughlin2016recurrent, gao2018revisiting, hermans2017defense,zheng2016mars} extract appearance features from each frame independently, and aggregate them into video-level representation by temporal pooling layer or recurrent neural network (RNN). In presence of partial occlusions, inaccurate detection and viewpoint variation, the learned features are often corrupted, result in significant performance degradation. Figure \ref{img1} illustrates some video sequences of pedestrians on MARS \cite{zheng2016mars} and iLIDS-VID \cite{wang2014person} datasets with these issues. Recent works attempt to address them by dividing video frames into horizontal rigid stripes \cite{fu2019sta, chen2020frame, wu2020adaptive} or utilizing attention mechanism \cite{li2020relation, li2018diversity, Subramaniam_2019_ICCV, ouyang2019video,hou2020temporal} to discover distinctive partial regions for extracting local appearance features. However, much background noise is blended in their located partial regions, thus they can not learn precise aligned part features from videos \cite{zhu2020identity}. Considering that, a few works \cite{chen2019spatial,jones2019body,gao2020pose,zhao2017spindle} employ pose estimation model \cite{miao2019pose} to adaptively locate key-points of pedestrians for extracting aligned part features. However, drastic viewpoint and pose variations as well as occlusion within videos affect the reliability of pose estimation model. Meanwhile, these methods only extract local features with fixed semantics from one-granularity partition, which can not cover all discriminative clues. Further, all the aforementioned methods only model the temporal relation across different frames, while neglecting complicated spatial-temporal dependencies and structural information of different body parts within a frame or across frames, restricting the capability of pedestrian representation.

In this work, we propose a novel Spatial-Temporal Correlation and Topology Learning framework (CTL) for video-based person re-identification, which pursues discriminative and robust representations. CTL extracts local features at multi-granularity levels to capture diverse discriminative semantics and alleviate unstable pose estimation results, and learns the potential cross-scale spatial-temporal dependencies and structure information among body parts for enhancing feature representation. Specifically, CTL employs a CNN backbone and a key-points estimator to extract semantic part features from human body at three granularities as graph nodes. It then explores a context-skeleton enriched topology to construct multi-scale graphs by considering both global contextual information and physical connections of human body, which effectively models the intrinsic spatial-temporal linkages between nodes. Moreover, a 3D graph convolution and a cross-scale graph convolution are designed for these multi-scale graphs, which facilitate direct cross-spacetime and cross-scale information propagation for capturing hierarchical spatial-temporal dependencies and structural information. By jointly performing the two convolutions, CTL effectively mines comprehensive and discriminative clues that are complementary with appearance information to enrich representation. Extensive experiments on two video datasets, \textit{i.e.}, MARS and iLIDS-VID, have demonstrated the effectiveness of the proposed approach.

Although graph modeling has been explored in person Re-ID, most of them only construct a graph on image-level \cite{shen2018person, chen2018group,yan2019learning,li2019adaptive} without considering temporal relation. A few of preliminary works \cite{wu2020adaptive, yan2020learning, yang2020spatial} extend graph modeling to video person Re-ID. However, they neglect the spatial structural information within each frame \cite{wu2020adaptive,yan2020learning}, or simply utilize factorized spatial and temporal graph modeling \cite{yang2020spatial}, failing to capture complex spatial-temporal relation. Further, all of them essentially belong to a local method. They utilize pair-wise feature affinity to measure the linkage between two nodes, while ignoring the impact of global contextual information from all other nodes, which is significance for learning reliable and useful graph topology.

The main contributions of this paper are as following: (1) We propose a novel Spatial-Temporal Correlation and Topology Learning framework (CTL) for person re-identification in videos. (2) We learn a context-reinforced topology to construct multi-scale graphs by considering both global contextual information and physical connections of human body. (3) We develop a 3D graph convolution and a cross-scale graph convolution to model high-oder spatial-temporal dependencies and structural information.

\section{Related Work}

\begin{figure*}[!t]
	\centering
	\includegraphics[width=6.9in]{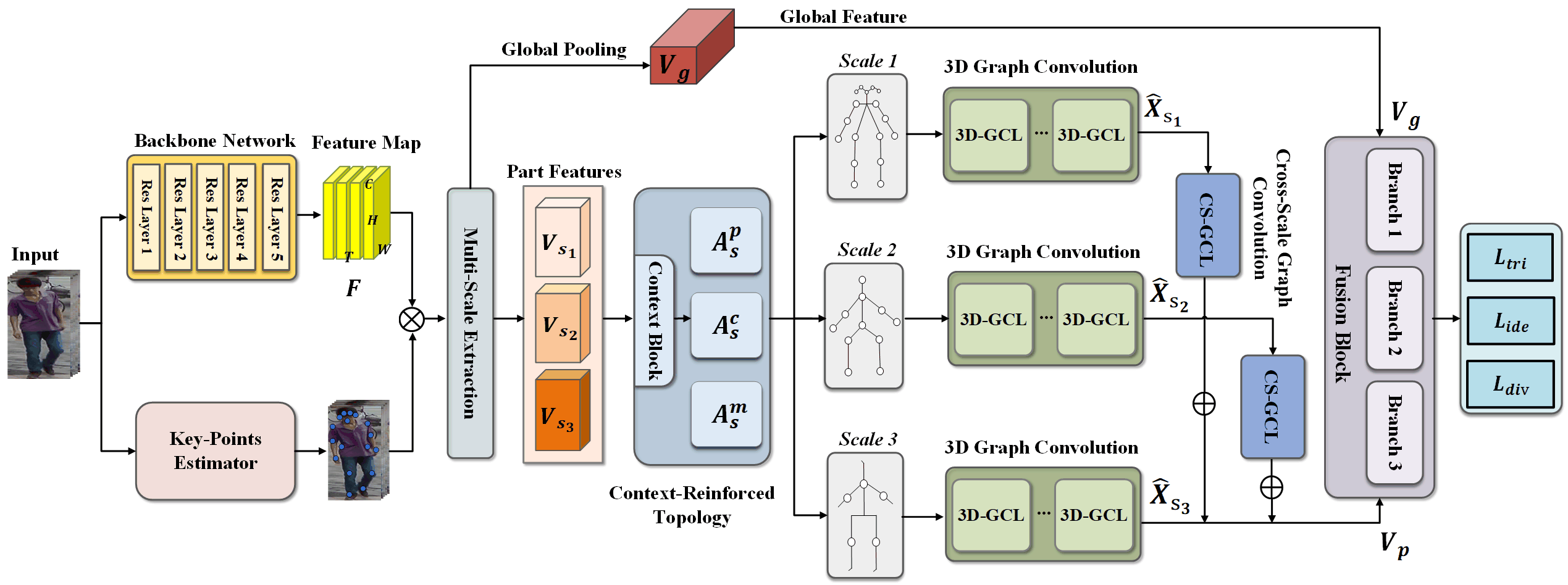}
	\caption{The overall architecture of the proposed CTL. It consists of a backbone network with a key-points estimator, a context block, multiple 3D graph convolution layers (3D-GCLs), multiple cross-scale graph convolution layers (CS-GCLs) and a fusion block.}
	\label{img2}
\end{figure*}

\textbf{Image-based Person Re-ID.} It is extensively explored in the literature. Existing methods mainly focus on three categories: designing discriminative hand-crafted descriptors \cite{chen2016similarity}, robust distance metric learning \cite{liao2015person,zhang2017improving} or deep learning technique \cite{liu2016multi,sun2018beyond,jiang2019rethinking,huang2019illumination,huang2020real}. For example, Chen \textit{et al.} \cite{chen2020salience}  introduced a cascaded feature suppression mechanism that mines all potential salient features stage-by-stage and integrates these discriminative salience features with the global feature, producing the final pedestrian feature.



\textbf{Video-based Person Re-ID.} Compared with image-based person Re-ID, video-based person Re-ID provides richer spatial-temporal clues and is promising for precise retrieval \cite{subramaniam2019co,hou2019vrstc,zhang2018learning}. Some existing works \cite{mclaughlin2016recurrent, gao2018revisiting, hermans2017defense} formulate video-based person Re-ID as an extension of image-based person Re-ID. They extract appearance representation from each frame, and aggregates the representations of all frames by using temporal pooling layer or RNN. For example, McLaughlin \textit{et al.} \cite{mclaughlin2016recurrent} proposed a siamese network, which captures features from each video, and then employs a recurrent layer and a temporal pooling layer to abstract video-level feature. In order to learn robust representation against partial occlusions, inaccurate detection and pose variation, rigid stripe partition \cite{fu2019sta, chen2020frame, wu2020adaptive} and attention mechanism \cite{li2020relation, li2018diversity, Subramaniam_2019_ICCV, ouyang2019video} methods attract more attention recently. For example, Subramaniam \textit{et al.} \cite{Subramaniam_2019_ICCV} formulated a Co-segmentation Activation Module to enhance common abstract features and suppress background features by jointly exploring common features across frames. Moreover, a few works \cite{chen2019spatial,jones2019body,gao2020pose} utilize pose estimation model to adaptively locate key-points of human body and learn aligned semantic features. For example, Jones \textit{et al.} \cite{jones2019body} proposed a pose-guided alignment framework, which mimicked the top-down attention of the human visual cortex to learn aligned features.

\textbf{Graph Learning.} Graphs are typically utilized to model relationships between nodes. Graph convolutional network (GCN) and its variant models \cite{kipf2016semi} have achieved great success in many computer vision tasks, \textit{e.g.}, object detection \cite{shi2020point},  multi-label image recognition \cite{chen2019multi} and skeleton-based action recognition \cite{zhang2020context, liu2020disentangling}. Similarity, some methods also apply GCN to person Re-ID. Most of them \cite{shen2018person, chen2018group,yan2019learning,li2019adaptive} build the graph models on image-level by considering the relations among images, which neglect the beneficial temporal information. In addition, a few of recent works \cite{wu2020adaptive, yan2020learning, yang2020spatial} extent GCN to video person Re-ID by exploring spatial and temporal relation, while they ignore the spatial structural information of body parts within each frame \cite{wu2020adaptive, yan2020learning} or only consider factorized spatial and temporal relation modeling \cite{yang2020spatial}. For example, Yang \textit{et al.} \cite{yang2020spatial} proposed a Spatial-Temporal Graph Convolutional Network (STGCN) which includes two GCN branches. The spatial branch learns spatial relation of human body, and the temporal branch mines discriminative temporal relation from adjacent frames.

\section{Method}
To further enhance the capacity of representations, this work explicitly explores spatial-temporal features across multi-granularity levels. To this end, we propose a Spatial-Temporal Correlation and Topology Learning framework, which models high-order spatial-temporal correlation to learn comprehensive representation. The overall architecture is shown in Figure \ref{img2}. It consists of a backbone network with a key-points estimator, a context block, multiple 3D graph convolution layers (3D-GCL), multiple cross-scale graph convolution layers (CS-GCL) and a fusion block. 


\subsection{Multi-Scale Feature Extraction}

\begin{figure}[!t]
	\begin{center}
		\includegraphics[width=0.85 \linewidth]{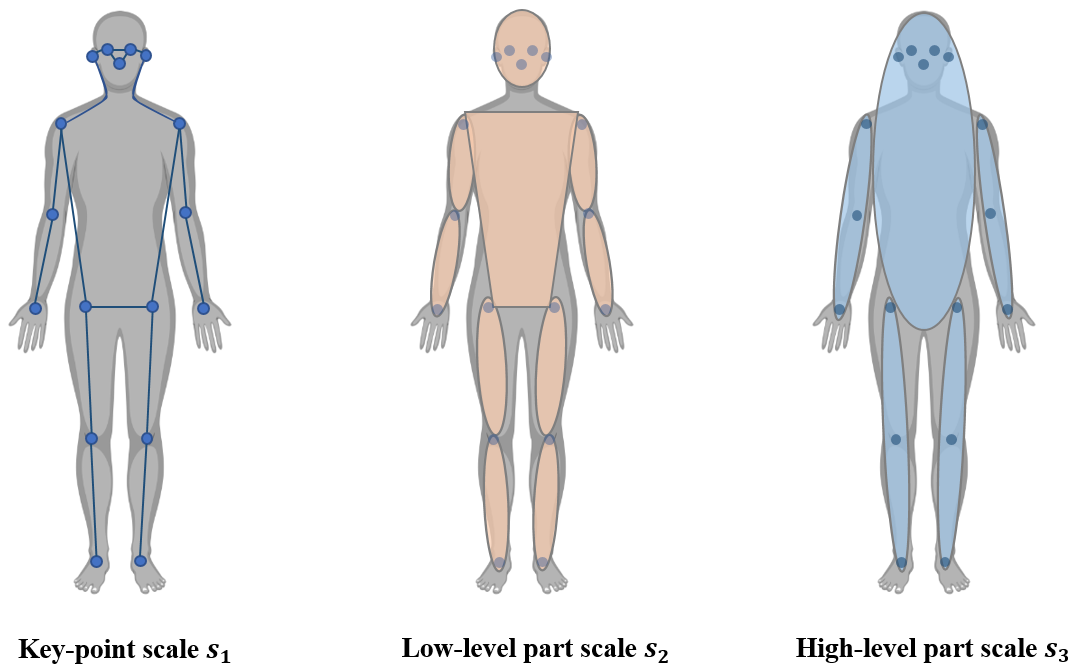}
	\end{center}
	\caption{Three scales of body partition. In $s_1$, we adopt 17 key-points, In $s_2$ and $s_3$, we
		adopt 10 and 5 parts of body, respectively.}
	\label{img3}
\end{figure}

Given a video sequence, we denote it as  $\{\boldsymbol{I}_t\}_{t=1}^T$, where $T$ is the sequence length. The backbone network takes each frame as input to extract the initial feature map $\boldsymbol{F}= \{\boldsymbol{F}_t | \boldsymbol{F}_t \in \mathbb{R}^{H\times W\times C} \}_{t=1}^T$, where $H$, $W$ and $C$ denote the height, width and channel size of the feature maps, respectively. The backbone network is based on ResNet-50 model \cite{he2016deep}. As part-based representations have shown effectiveness for person Re-ID \cite{sun2018beyond}, we adopt a key-points estimator \cite{sun2019deep} to adaptively locate the key-points of human body, and extract aligned part features from these key-points against partial occlusions, misalignment and viewpoint variation. Although, key-points estimation models have obtained high accuracy, they remain suffering from unreliable performance under complex surveillance scenes, leading to inaccurate key-points location and their confidence. Thus, exploring multi-scale part features with their spatial-temporal correlation is particularly important, which can alleviate unreliable key-point estimation results and capture diverse discriminative semantics. 

Based on human nature, we divide human body at three granularities: the key-point scale ($s_1$), the low-level-part scale ($s_2$) and the high-level-part scale ($s_3$), as show in Figure \ref{img3}. We merge spatially nearby key-points to each part in coarser scales based on human prior. The heat maps $\boldsymbol{m}$ of key-points are generated through the key-point estimator, which are then normalized with a softmax function. The group of semantic local feature $\boldsymbol{V}_{s1} \in \mathbb{R}^{T\times N_{s_1}\times C}$ for the granularity $s_1$ and the global feature $\boldsymbol{V}_{g} \in \mathbb{R}^{T\times C}$ are computed as following:
\begin{equation}
\begin{split}
\boldsymbol{V}_{s1} & = \{\boldsymbol{v}_{s1}\} = g_{GAP}(\boldsymbol{F}_t\otimes\boldsymbol{m}_{s_1}^t) \\
\boldsymbol{V}_{g} & = g_{GAP}(\boldsymbol{F}) 
\end{split}
\end{equation}
where $\otimes$ and $g_{GAP}$ refer to outer product and global average pooling operations, respectively. $N_s$ is the number of body parts (\textit{17, 10 and 5 parts for $s_1$, $s_2$, $s_3$, respectively}). The part features $\boldsymbol{V}_{s2}$ and $\boldsymbol{V}_{s3}$ for the low-level-part and high-level-part scales are computed by performing average pooling operation on the features $\boldsymbol{V}_{s1}$ of the key-points within each body part.

\subsection{Context-Reinforced Topology Graph}
In order to excavate spatial-temporal information from video frames, we employ advanced GCN to model hierarchical spatial-temporal dependencies and structural information. Let $\mathcal{G} = \{\mathcal{G}_s\}_{s\in\{s_1,s_2,s_3\}}$ be  a set of constructed multi-scale graphs of one video frame, where each graph corresponds to a specific granularity level $s$. Specifically, $\mathcal{G}_s(\mathcal{V}_s, \mathcal{E}_s)$ includes $N_s$ nodes $\boldsymbol{v}_i \in \mathcal{V}_s$ and a set of edges $\boldsymbol{e}_{ij} = (\boldsymbol{v}_i, \boldsymbol{v}_j) \in \mathcal{E}_s$. Each part of body within one video frame is viewed as a graph node and the edges represent the relationship between these body parts. The input node feature of frame $t$ is denoted as $\boldsymbol{X}_s^t = \boldsymbol{V}_{s}^t \in \mathbb{R}^{N_s \times C} $. $\boldsymbol{A}_s \in \mathbb{R}^{N_s\times N_s}$ is the corresponding frame-level adjacency matrix, in which each element represents the linkage of two arbitrary nodes. The topology of the graph is actually decided by $\boldsymbol{A}_s$. Existing GCN-based Re-ID methods predict the relationship between two nodes independently by calculating pair-wise feature affinity, which ignore the impact of all other contextual nodes and only consider undirected dependency, restricting the capacity and expressiveness of the graph model.

\begin{figure}[!t]
	\begin{center}
		\includegraphics[width=1.0 \linewidth]{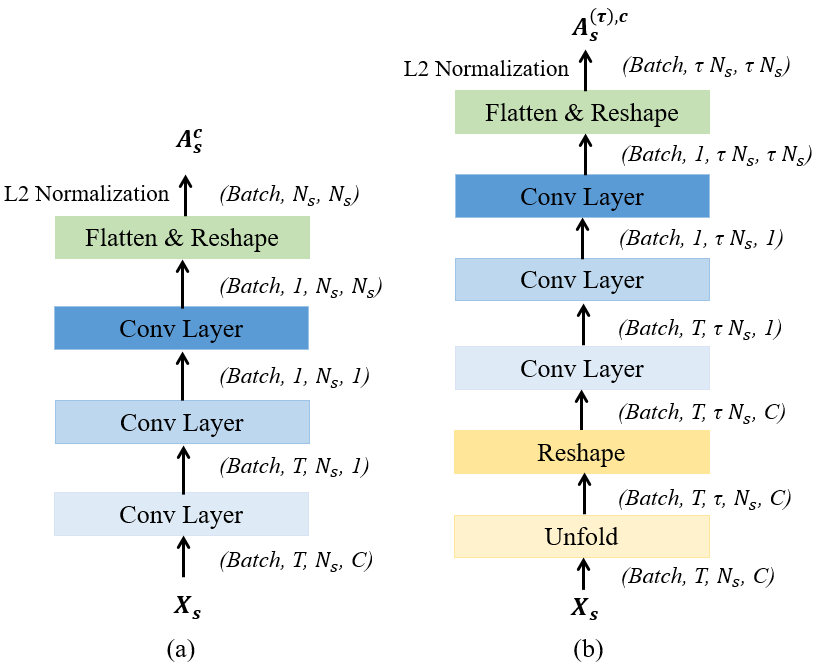}
	\end{center}
	\caption{Detailed network structure of (a) the context block; (b) the advanced context block.}
	\label{img4}
\end{figure} 

Considering that, we explore a context-reinforced topology to construct graph, which simultaneously encodes contextual information along the node, temporal and feature dimensions, as well as physical structural information of human body. The context-reinforced adjacency matrix $\boldsymbol{A}_s$ consists of three components:
\begin{equation}
\begin{split}
\boldsymbol{A}_{s} = \boldsymbol{A}_s^p + \boldsymbol{A}_s^m + \boldsymbol{A}_s^c
\end{split}
\end{equation}
where $\boldsymbol{A}_s^p \in \{0,1\}^{N_s\times N_s}$ denotes the physical connections of human body with rich structural information, which is fixed during training. $\boldsymbol{A}_s^m$ denotes a mask adjacency matrix, which is utilized as the attention on the physical structure, inspired by \cite{shi2019skeleton}. $\boldsymbol{A}_s^m$ improves the flexibility and generality of static global graph structure $\boldsymbol{A}_s^p$, and is initialized with zeros and optimized together with other parameters during training. $\boldsymbol{A}_s^c$ is a data-dependent individual adjacency matrix, which incorporates the global contextual information of all nodes and learns a unique dynamic topology graph for each sample. $\boldsymbol{A}_s^c$ is learned by a context block, as shown in Figure \ref{img4}(a). Given the node features $\{\boldsymbol{X}_{s}^t\}_{t=1}^T \in  \mathbb{R}^{T\times N_s \times C}$, the context block firstly squeezes the feature and temporal dimensions of each node by two convolution layers with $1\times1$ kernel. Then, it utilizes an addition $1\times 1$ convolution layer to transfer the $N_s$-dimension feature vector into the $N_s \times N_s$ adjacency matrix $\boldsymbol{A}_s^c$. Afterwards, $L2$ normalization operation is applied to each row of $\boldsymbol{A}_s^c$ for stable optimization. The context block adequately considers the influence of all other nodes when measuring the relationship between two arbitrary nodes. 

\subsection{3D Graph Convolutional Layer}

\begin{figure}[!t]
	\begin{center}
		\includegraphics[width=0.9 \linewidth]{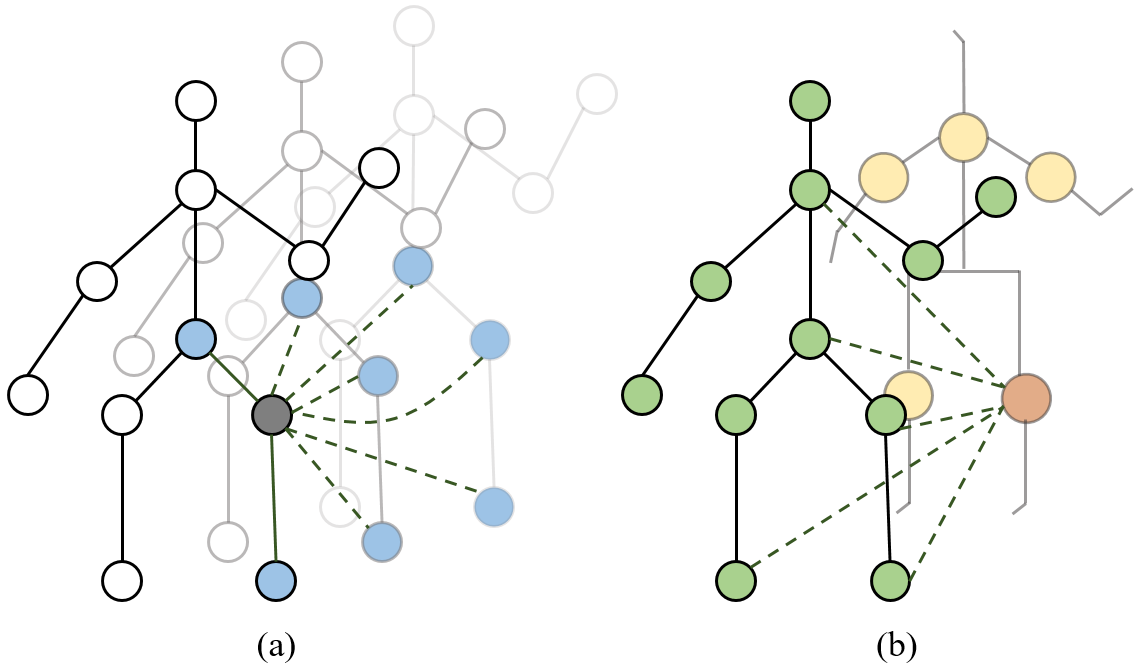}
	\end{center}
	\caption{(a) Cross-spacetime information propagation by 3D graph convolution; (b) Cross-scale information propagation by cross-scale graph convolution.}
	\label{img5}
\end{figure}
After obtaining the frame-level graphs for all frames, we design a 3D graph convolution to effectively propagate messages and update node features. 3D-GCL allows direct cross-spacetime information propagation for capturing complex spatial-temporal dependencies and structural information in a spatial-temporal graph, as shown in Figure \ref{img5}(a). Concretely, 3D-GCL first uses a temporal sliding window with size of $\tau$ over the sequence of frame-level graphs. At each sliding step, a spatial-temporal subgraph $\mathcal{G}_s^{(\tau)} = (\mathcal{V}^{(\tau)}_s, \mathcal{E}^{(\tau)}_s)$, in which $\mathcal{V}^{(\tau)}_s = \mathcal{V}^1_s \cup ... \cup \mathcal{V}^{\tau}_s$ denote the union set of all nodes across $\tau$ video frames in this window. And the edge set $\mathcal{E}^{(\tau)}_s$  is represented by a block adjacency matrix $\boldsymbol{A}_s^{(\tau)}$. It is computed as following:
\begin{equation}
\begin{split}
\boldsymbol{A}_s^{(\tau)} &= \boldsymbol{A}_s^{(\tau),p} + \boldsymbol{A}_s^{(\tau),m} +\boldsymbol{A}_s^{(\tau),c}\\
&=\begin{bmatrix}
	[\boldsymbol{A}_s^{(\tau)}]_{1,1}   & \cdots   & [\boldsymbol{A}_s^{(\tau)}]_{1,\tau}   \\
	\vdots    & \ddots   & \vdots  \\
	[\boldsymbol{A}_s^{(\tau)}]_{\tau,1}  & \cdots\  & [\boldsymbol{A}_s^{(\tau)}]_{\tau,\tau}  \\
\end{bmatrix}  
\in\mathbb{R}^{\tau N_s \times \tau N_s}
\end{split}
\end{equation}
where each submatrix $[\boldsymbol{A}_s^{(\tau)}]_{i,j}$ denotes the graph nodes of $\mathcal{V}^i_s$ are connected to themselves and their temporal neighboring nodes at frame $j$, by expanding the frame-level spatial connection (corresponding to $[\boldsymbol{A}_s^{(\tau)}]_{i,i}$) to the temporal domain. $\boldsymbol{A}_s^{(\tau)}$ is still composed of three parts. The block adjacency matrix $\boldsymbol{A}_s^{(\tau),p}$ is computed by tiling the static $\boldsymbol{A}_s^p$ in each block. $\boldsymbol{A}_s^{(\tau),m}$ is obtained as the same way. $\boldsymbol{A}_s^{(\tau),c}$ is learned by the advanced context block in Figure \ref{img4}(b). Simultaneously, $\boldsymbol{X}_s^{(\tau)} \in \mathbb{R}^{T\times \tau N_s\times C}$ is obtained by employing the sliding temporal window over $\boldsymbol{X}^0 = \{\boldsymbol{X}_s^t\}_{t=1}^T \in \mathbb{R}^{T\times N_s\times C}$ with zero padding operation to build $T$ windows, which is the input of 3D-GCL. 

The 3D graph convolution for the $t$-th temporal window at $l$-th iteration is formulated as following:
\begin{equation}
\begin{split}
[\boldsymbol{X}_s^{(\tau),l+1}]_t = \sigma( \tilde{\boldsymbol{D}}^{-\frac{1}{2}}\boldsymbol{\hat{A}}_{s,t}^{(\tau)}\tilde{\boldsymbol{D}}^{-\frac{1}{2}}[\boldsymbol{X}_s^{(\tau),l}]_t\boldsymbol{W}^l)
\end{split}
\end{equation}
where $\small(\tilde{\boldsymbol{D}})_{i,i} = \sum_j (\boldsymbol{\hat{A}}_{s,t}^{(\tau)})_{i,j}$ denotes diagonal node degree matrix \cite{kipf2016semi}, $\small\boldsymbol{\hat{A}}_{s,t}^{(\tau)} = \boldsymbol{A}_{s,t}^{(\tau)} + I_{\tau N_s}$ denotes the self-loop adjacency matrix. $\boldsymbol{I}_{\tau N_s}$ denotes an identity matrix. $\boldsymbol{W}^l$ refers to the learnable parameter and $\sigma$ represents a non-linear activation function. After each 3D-GCL, a convolution layer followed with a batch normalization (BN) layer and a rectified linear units (ReLU) layer is employed to collapse the window dimension $\tau$ and output the updated node feature $\boldsymbol{X}^{l+1} \in \mathbb{R}^{T\times N_s \times C}$ . In addition, shortcut connection $\boldsymbol{X}^{l+1} = \boldsymbol{X}^{l+1} + \boldsymbol{X}^{l}, 1\leq l \leq L-1$ is adopted for effective and stable optimizing. The refined part features at three granularities  $\boldsymbol{\hat{X}}_{s_1}$, $\boldsymbol{\hat{X}}_{s_2}$, $\boldsymbol{\hat{X}}_{s_3}$ are finally obtained by performing multiple 3D-GCLs.  

\subsection{Cross-Scale Graph Convolutional Layer}

\begin{figure}[!t]
	\begin{center}
		\includegraphics[width=1.0 \linewidth]{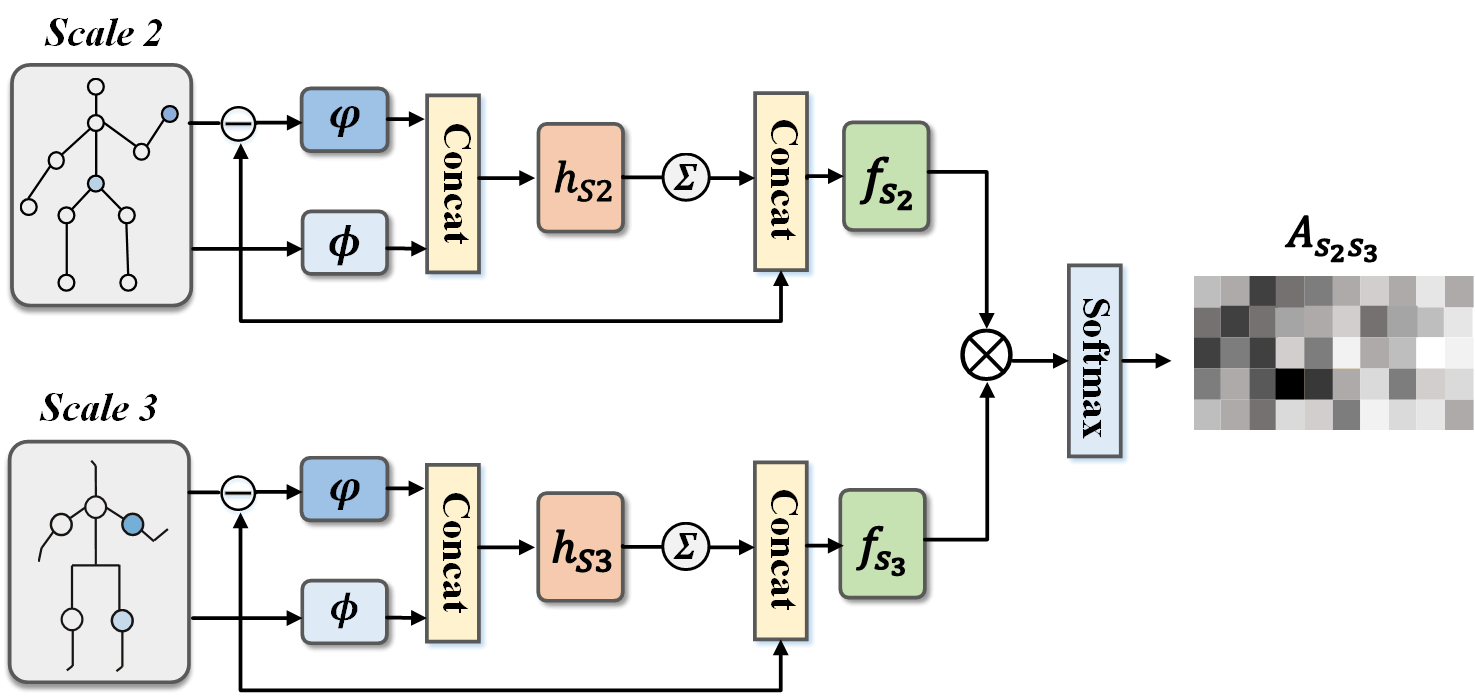}
	\end{center}
	\caption{The inference of the cross-scale adjacent matrix.}
	\label{img6}
\end{figure}

Multi-scale part features obtained from different partitions contain diverse discriminative semantics. To enable information diffusion across scales and learn comprehensive representation, we propose a cross-scale graph convolution, which propagates the informative clues of part features from one scale to another, as shown in Figure \ref{img5}(b). The cross-scale topology graph is a directed graph that corresponds the nodes in one scale graph to the nodes in another scale graph. For simplicity, we elaborate a CS-GCL associated from $s_2$ to $s_3$. The adjacent matrix $\boldsymbol{A}_{s_2,s_3} \in \mathbb{R}^{N_{s_3} \times N_{s_2}}$ of the cross-scale graph predicts the cross-scale relationship. As shown in Figure \ref{img6}, the dependency $(\boldsymbol{A}_{s_2,s_3})_{i,m}$ between $i$-th part in $s_2$ and $m$-th part in $s_3$ is computed as following:
\begin{equation}
\begin{split}
&\boldsymbol{p}_{i,s_2} = \sum_{j=1}^{N_{s_2}} h_{s_2}([{\phi} (\boldsymbol{x}_{i,s_2}), \varphi (\boldsymbol{x}_{j,s_2}- \boldsymbol{x}_{i,s_2})]) \\
&\boldsymbol{r}_{i,s_2} = f_{s_2}([\boldsymbol{x}_{i,s_2}, \boldsymbol{p}_{i,s_2}])\\
&\boldsymbol{p}_{m,s_3} = \sum_{j=1}^{N_{s_3}} h_{s_3}([{\phi} (\boldsymbol{x}_{m,s_3}), \varphi (\boldsymbol{x}_{j,s_3}- \boldsymbol{x}_{m,s_3})]) \\
&\boldsymbol{r}_{m,s_3} = f_{s_3}([\boldsymbol{x}_{m,s_3}, \boldsymbol{p}_{m,s_3}])\\
&(\boldsymbol{A}_{s_2,s_3})_{i,m}  = softmax(\boldsymbol{r}_{m,s_3}^\top \boldsymbol{r}_{i,s_2})
\end{split}
\end{equation}
where $\boldsymbol{x}_{i,s_2} \in \mathbb{R}^C $ denotes the $i$-th component of  $\boldsymbol{\hat{X}}_{s_2}$ at one specific frame. $h_{s_2}$, $f_{s_2}$, $\phi$, $\varphi$ are the embedding functions implemented by a full connected layer with a BN layer and a ReLU layer. $\boldsymbol{p}_{i,s_2}$ and $\boldsymbol{p}_{m,s_3}$ aggregate the global relation information of all other part features to the $i$-th and the $m$-th components at the two scales. $\boldsymbol{r}_{i,s_2}$ and $\boldsymbol{r}_{m,s_3}$ are the augmented global relation features, which are then used to calculate the dependency $(\boldsymbol{A}_{s_2,s_3})_{i,m}$ by inner product operation and softmax function. Thus, $\boldsymbol{A}_{s_2,s_3}$ constructs the influence from the body in $s_2$ to each part in $s_3$. 

Given the part feature  $\boldsymbol{\hat{X}}_{s_2}$ at scale $s_2$, the cross-scale convolution for frame $t$ is formulated as following:
\begin{equation}
\begin{split}
[\boldsymbol{\hat{X}}_{s_{23}}]_t =\sigma(\boldsymbol{A}_{s_2,s_3}^t [\boldsymbol{\hat{X}}_{s_2}]_t\boldsymbol{W}_{s_{23}})
\end{split}
\end{equation}
where $\boldsymbol{W}_{s_{23}}$ denotes the parameter matrix, and $\boldsymbol{\hat{X}}_{s_{23}} $ is the transformed part feature. Such feature adaptively absorbs informative clues from the corresponding parts of body in $s_2$. Analogously, we also utilize another CS-GCL to transfer the part feature $\boldsymbol{\hat{X}}_{s_1}$ from $s_1$ to $s_3$, and produce the transformed part feature $\boldsymbol{\hat{X}}_{s_{13}}$. Finally, the comprehensive part feature $\boldsymbol{V}_p$ with three-granularity information is obtained as following:
\begin{equation}
\begin{split}
\boldsymbol{V}_p = \boldsymbol{\hat{X}}_{s_3} + \alpha(\boldsymbol{\hat{X}}_{s_{13}} + \boldsymbol{\hat{X}}_{s_{23}})
\end{split}
\label{eq1}
\end{equation}
where $\alpha$ denotes the balance weight.

\subsection{Model Optimizing}
After obtaining the features $\boldsymbol{V}_p \in \mathbb{R}^{T\times N_{s_3}\times C} $ and $\boldsymbol{V}_g$, they are fed into a fusion block to further incorporate the global and local information, and are finally optimized by loss function. The fusion block consists of three branches. The first branch employs a temporal average pooling layer ($g_{TAP}$) for $\boldsymbol{V}_g$ to generate the feature vector $\small\boldsymbol{V}_f^g = g_{TAP}(\boldsymbol{V}_g)$. The second branch utilizes the function $\small g_{TAP}((\sum\limits_{n=1}^{N_{s_3}} [\boldsymbol{V}_p]_{:,n,:}) + \boldsymbol{V}_g)$ to generate the feature vector $\boldsymbol{V}_f^a$. The third branch utilizes the function $\small g_{TAP}([g_c([\boldsymbol{V}_p]_{:,1,:}),...,g_c([\boldsymbol{V}_p]_{:,N_{s_3},:})] + \boldsymbol{V}_g)$ to produce the feature vector $\boldsymbol{V}_f^c$, where $g_c$ denotes a $1\times 1$ convolution layer for reducing the dimension. The third branch implicitly promotes the channel-wise semantic alignment between the global and local features, which drives different channel of
the global feature to focus on different body parts for improving the performance. Identification loss and triplet loss are the widely-used losses for person re-identification, we adopt triplet loss with hard mining strategy \cite{yang2020spatial} and identification loss with label smoothing regularization \cite{szegedy2016rethinking} to optimize these three features $\boldsymbol{V}_f^g$, $\boldsymbol{V}_f^a$ and $\boldsymbol{V}_f^c$, respectively. The two losses are denoted as $\mathcal{L}_{tri}$ and $\mathcal{L}_{ide}$ respectively. Moreover, a diversity regularization loss is proposed to encourage the diversity of the local features and increase the discrimination of the final video representation. This loss is defined as following:
\begin{equation}
\begin{split}
\mathcal{L}_{div} = \lVert \boldsymbol{V}_p\boldsymbol{V}_p^T- \boldsymbol{I} \rVert^2_F
\end{split}
\end{equation}
where $\lVert \cdot \lVert_F$ denotes Frobenius norm. $\boldsymbol{V}_p$ is applied with temporal average pooling and $L2$ normalization in advance for this loss. Therefore, the total loss $\mathcal{L}$ for CTL is the combination of the three losses:
\begin{equation}
\begin{split}
\mathcal{L}= \lambda_1 \cdot \mathcal{L}_{tri} + \lambda_2 \cdot \mathcal{L}_{ide} + \lambda_3 \cdot \mathcal{L}_{div}
\end{split}
\label{eq2}
\end{equation}
where $\lambda_{1-3}$ are the balance weights of the three loss terms.

\section{Experiments}




\subsection{Experimental Settings}
\textbf{Datasets.} MARS dataset \cite{zheng2016mars} is one of the existing largest video benchmark, consisting of 1,261 identities and 20,715 video sequences. The training set contains 625 identities and the testing set contains 636 identities. iLIDS-VID dataset \cite{wang2014person} is a small-scale benchmark. It consists of 600 video sequences of 300 different identities, each of which has two sequences captured by two non-overlapping cameras. It is randomly split into a training set with 150 identities and a testing set with the remaining 150 identities.

\textbf{Evaluation Metrics.} We adopt the standard metrics, \textit{i.e.}, Cumulative Matching Characteristic (CMC) curves and mean average precision (mAP), to evaluate the performance of different person Re-ID algorithms.

\textbf{Implementation Details.} We randomly sample $T = 6$ frames from a variable-length sequence as a input clip. Each mini-batch has 8 identities and 4 video clips for each identity. We resize all video frames to  $256 \times 128$ pixels, which are normalized with $1.0/256$. We then apply image-level data augmentation to each video clip, including random horizontal flipping and random erasing probability. ResNet-50 \cite{he2016deep} pre-trained on ImageNet is used as the backbone network. The last stride of ResNet-50 is set to 1. The Adam optimizer is adopted with the initial learning rate $lr$ of $3e^{-4}$ and the weight decay of $5e^{-4}$. We train our model for 240 epochs in total. The learning rate $lr$ is decayed by 10 after every 60 epochs. $H$, $W$ and $C$ are 16, 8 and 2048, respectively. The numbers of 3D-GCLs is set to $L=2$. $\alpha$ in Eq.~\ref{eq1} is set to $0.3$, and  $\lambda_{1-3}$ in Eq.~\ref{eq2} are all set to 1. During inference, $\boldsymbol{V}_f^a$ is used as the final video representation for calculating the similar scores. 

\subsection{Comparison to State-of-the-Arts}
\begin{table}[!t]
	\scriptsize
	\caption{Performance comparison to the state-of-the-art methods on MARS dataset.}
	\begin{center}
		\newcommand{\tabincell}[2]{\begin{tabular}{@{}#1@{}}#2\end{tabular}}
		\resizebox{\columnwidth}{!}{
			\renewcommand{\arraystretch}{1.3}
			\begin{tabular}{c|c|c|c|c}
				\hline
				\textbf{Method}&\textbf{Rank-1}&\textbf{Rank-5}&\textbf{Rank-20}&\textbf{mAP}\\
				\hline
				CNN+XQDA \cite{zheng2016mars}&68.3&82.6&89.4&49.3\\
			    TriNet \cite{hermans2017defense}& 79.8& 91.36 &-& 67.7\\
				\hline
				STAL\cite{chen2019spatial} &82.2& 92.8& 98.0 &73.5\\
				STAN \cite{li2018diversity}&82.3&-&-&65.8\\
				COSAM\cite{subramaniam2019co}&84.9&95.5&97.9&79.9\\
				VRSTC \cite{hou2019vrstc}&88.5&96.5&97.4&82.3\\
				RGSAT \cite{li2020relation}&89.4&96.9&98.3&84.0\\
				AGRL \cite{wu2020adaptive}&89.8 &96.1& 97.6 &81.1\\
				TCLNet \cite{hou2020temporal}&89.8&-&-&85.1\\
				STGCN \cite{yang2020spatial} &90.0&96.4&98.3&83.7\\
				MGH \cite{yan2020learning} & 90.0& 96.7 &98.5& 85.8\\
				AP3D\cite{gu2020appearance} &90.1&-&- &85.1\\
				AFA \cite{chen2020temporal}  &90.2& 96.6&-& 82.9\\
				\hline 
				CTL &\textbf{91.4}&\textbf{96.8}&\textbf{98.5}&\textbf{86.7}\\
				\hline
		\end{tabular}}
	\end{center}
	\label{table1}
\end{table}

\begin{table}[!t]
	\tiny
	\caption{Performance comparison to the state-of-the-art methods on iLIDS-VID dataset.}
	\begin{center}
		\newcommand{\tabincell}[2]{\begin{tabular}{@{}#1@{}}#2\end{tabular}}
		\resizebox{\columnwidth}{!}{
			\renewcommand{\arraystretch}{1.2}
			\begin{tabular}{c|c|c|c}
				\hline
				\textbf{Method}&\textbf{Rank-1}&\textbf{Rank-5}&\textbf{Rank-20}\\
				\hline
				CNN+XQDA \cite{zheng2016mars}&53.0&81.4&95.1\\
				RCNet \cite{mclaughlin2016recurrent}&58& 84.0& 96.0\\
				\hline
				COSAM\cite{subramaniam2019co}&79.6&95.3&-\\
				STAN \cite{li2018diversity}&80.2&-&-\\
				STAL\cite{chen2019spatial} &82.8 &95.3& 98.8\\
				VRSTC \cite{hou2019vrstc}&83.4&95.5&99.5\\
				AGRL \cite{wu2020adaptive}&83.7 &95.4& 99.5\\
			    MGH \cite{yan2020learning} & 85.6 &97.1 &99.5\\
				RGSAT \cite{li2020relation}&86.0& \textbf{98.0}&99.4\\
				TCLNet \cite{hou2020temporal}&86.6&-&-\\
				AP3D\cite{gu2020appearance} &86.7&-&-\\
				FGRA \cite{chen2020frame} &88.0 &96.7 & 99.3\\
				AFA \cite{chen2020temporal} &88.5& 96.8 &99.7\\
				\hline 
				CTL &\textbf{89.7}&97.0&\textbf{100.0}\\
				\hline
		\end{tabular}}
	\end{center}
	\label{table2}
\end{table}

\textbf{Results on MARS.}  In Table \ref{table1}, we compare the proposed method with 13 state-of-the-art methods on MARS dataset. The first two methods belong to the expansion of image-based person Re-ID. We can observe that CTL achieves 91.4\% Rank-$1$ accuracy and 86.7\% mAP, surpassing the current state-of-the-art methods by a large margin. It improves the 2nd best method AFA \cite{chen2020temporal} by 1.2\% Rank-$1$ accuracy and MGH \cite{yan2020learning} by 0.9\% mAP, respectively. The comparison clearly demonstrates the effectiveness and superiority of CTL for exploring hierarchical spatial-temporal dependencies and structural information among body parts from videos. Note that, compared with other graph-based methods, including AGRL\cite{wu2020adaptive}, STGCN \cite{yang2020spatial} and MGH \cite{yan2020learning}, CTL obtains better results in terms of Rank-$1$ accuracy and mAP. The main reason for the boosting is two aspects: 1) using context-reinforced topology to construct graph instead of pair-wise feature affinity; 2) the advantage of modeling high-order spatial-temporal correlation by 3D-GCL and CS-GCL. 

\textbf{Results on iLIDS-VID.}  Table \ref{table2} reports the performance of our approach with 13 state-of-the-art methods on iLIDS-VID dataset. CNN+XQDA \cite{zheng2016mars} and RCNet \cite{mclaughlin2016recurrent} are the straightforward expansion method of image-based person Re-ID. It can be seen that CTL obtains the best performance of  89.7\% Rank-$1$ accuracy and 100.0\% Rank-$20$ accuracy. It beats AFA \cite{chen2020temporal} on Rank-$1$ and Rank-$20$ accuracy by 1.2\% and by 0.3\%, respectively. The comparison demonstrates the advantage of spatial-temporal feature learning by CTL, and the applicability of CTL for a small-scale dataset.

\subsection{Ablation Studies}

\textbf{Effectiveness of Components.} Table \ref{table3} reports the experimental results of the ablation studies for CTL. Basel denotes using the backbone network with the key-points estimator to learn the global and multi-scale part features. Basel+ContRe denotes using the context-reinforced topology to structure multi-scale frame-wise graphs, and applying original GCN to learn the refined multi-scale part features and global feature. Basel+ContRe+3D refers to CTL using addition operation to replace CS-GCL for learning the fused part feature and the global feature.  Basel+ContRe+3D+CS refers to the whole framework of CTL. Compared with Basel, Basel+ContRe boosts Rank-$1$ accuracy and mAP by 0.8\% and 2.7\%. This indicts that the effectiveness of the context-reinforced topology to capture the intrinsic relationship among body parts for enhancing feature representation. Moreover, Basel+ContRe+3D improves Basel+ContRe by 1.3\% Rank-1 accuracy and 0.6\% mAP, which verifies the effectiveness of 3D-GCL for allowing direct cross-spacetime information propagation to enrich part features. By utilizing the cross-scale graph convolutional layer, Basel+ContRe+3D+CS achieves the best performance. This demonstrates that CS-GCL effectively capture diverse visual semantic across multiple scales to integrate them into a comprehensive representation.

\textbf{Analysis of Context-Reinforced Topology.} The results in Table \ref{table4} show the influence of different components of the context-reinforced graph topology. CTL-$\boldsymbol{A}_s^p$, CTL-$\boldsymbol{A}_s^{p}\boldsymbol{A}_s^{m}$, CTL-$\boldsymbol{A}_s^{p}\boldsymbol{A}_s^{m}\boldsymbol{A}_s^{c}$, denote using $\boldsymbol{A}_s^{m}$,  $\boldsymbol{A}_s^{p}+\boldsymbol{A}_s^{m}$ and $\boldsymbol{A}_s^{p}+\boldsymbol{A}_s^{m}+\boldsymbol{A}_s^{c}$ to measure the dependency between two nodes, respectively. By comparing CTL-$\boldsymbol{A}_s^{p}$ and CTL-$\boldsymbol{A}_s^{p}\boldsymbol{A}_s^{m}$, we can conclude that $\boldsymbol{A}_s^{m}$ improves the flexibility of graph topology and captures more complex spatial-temporal correlation. CTL-$\boldsymbol{A}_s^{p}\boldsymbol{A}_s^{m}\boldsymbol{A}_s^{c}$ achieves remarkable performance improvement as compared to CTL-$\boldsymbol{A}_s^{p}\boldsymbol{A}_s^{m}$, which means $\boldsymbol{A}_s^{c}$ is complementary to physical topology, and mines potential connections that are informative by considering global contextual information of all nodes.

\begin{figure}[!t]
	\begin{center}
		\includegraphics[width=0.95 \linewidth]{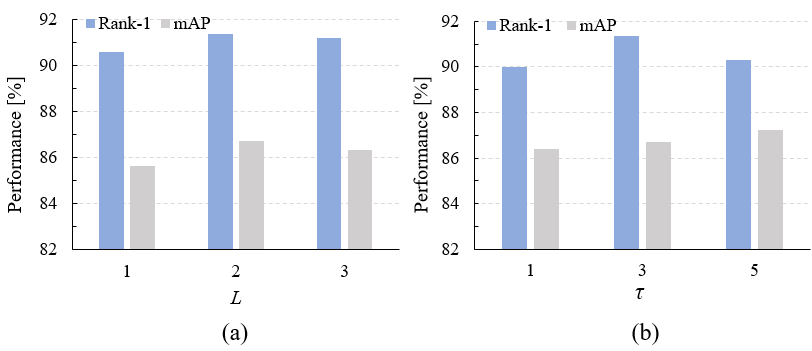}
	\end{center}
	\caption{Analysis on the influence of different hyperparameters for 3D-GCL, (a) the number layer $L$; (b) windows size $\tau$.}
	\label{img7}
\end{figure}

\textbf{Analysis of 3D-GCL.} We conduct the experiments to analyze the influence of the number layer $L$ and window size $\tau$ for 3D-GCL. In Figure \ref{img7}(a), we can observe that the best Rank-1 accuracy of 91.4\% and mAP of 86.7\% are obtained when $L=2$. This implies that one-layer 3D-GCL has insufficient capability for capturing complex spatial-temporal information, whilst three-layer 3D-GCLs bring more training parameters, result in hard optimizing and performance degradation. Thus, $L$ is set to $2$. In Figure \ref{img7}(b), $\tau=3$ obtains superior performance as compared to $\tau=1$ due to utilizing the temporal complementary information from local temporal neighborhood nodes, but the gain diminishes when $\tau = 5$ as the discriminative clues in aggregated features are counteracted due to the over-sized local temporal neighborhood nodes. Thus, $\tau$ is set to $3$.

\begin{table}[!t]
	\caption{Evaluation of the effectiveness of each component of
		CTL on MARS dataset.}
	\begin{center}
		\newcommand{\tabincell}[2]{\begin{tabular}{@{}#1@{}}#2\end{tabular}}
		\resizebox{\columnwidth}{!}{
			\renewcommand{\arraystretch}{1.4}
			\begin{tabular}{c|c|c|c|c}
				\hline
				\textbf{Model}&\textbf{Rank-1}&\textbf{Rank-5}&\textbf{Rank-20}&\textbf{mAP}\\
				\hline
				Basel&88.6&96.1&97.9&82.7\\
				Basel+ContRe&89.4&95.6&98.2&85.4\\
				Basel+ContRe+3D&90.7&96.4&98.4&86.0\\
				Basel+ContRe+3D+CS&91.4&96.8&98.5&86.7\\
				\hline
		\end{tabular}}
	\end{center}
	\label{table3}
\end{table}

\begin{table}[!t]
	\tiny
	\caption{Evaluation of the influence of different components of the context-reinforced topology on MARS dataset.}
	\begin{center}
		\newcommand{\tabincell}[2]{\begin{tabular}{@{}#1@{}}#2\end{tabular}}
		\resizebox{\columnwidth}{!}{
			\renewcommand{\arraystretch}{1.4}
			\begin{tabular}{c|c|c|c|c}
				\hline
				\textbf{Model}&\textbf{Rank-1}&\textbf{Rank-5}&\textbf{Rank-20}&\textbf{mAP}\\
				\hline
				CTL-$\boldsymbol{A}_s^p$   &90.4&96.4&98.5&86.1\\
				CTL-$\boldsymbol{A}_s^{p}\boldsymbol{A}_s^{m}$ &90.9&96.3&98.6&86.3\\
				CTL-$\boldsymbol{A}_s^{p}\boldsymbol{A}_s^{m}\boldsymbol{A}_s^{c}$ &91.4&96.8&98.5&86.7\\
				\hline
		\end{tabular}}
	\end{center}
	\label{table4}
\end{table}

\begin{table}[!t]
	\tiny
	\caption{Evaluation of the influence of CS-GCL with different settings
		on MARS dataset.}
	\begin{center}
		\newcommand{\tabincell}[2]{\begin{tabular}{@{}#1@{}}#2\end{tabular}}
		\resizebox{\columnwidth}{!}{
			\renewcommand{\arraystretch}{1.4}
			\begin{tabular}{c|c|c|c|c}
				\hline
				\textbf{Model}&\textbf{Rank-1}&\textbf{Rank-5}&\textbf{Rank-20}&\textbf{mAP}\\
				\hline
				CS-GCL($\boldsymbol{M}=1$) &91.4&96.8&98.5&86.7\\
				CS-GCL($\boldsymbol{M}=2$) &90.1&96.5&98.3&85.5\\\hline
				CS-GCL-$\boldsymbol{s}_3$ &90.4&96.4&98.4&85.9\\
				CS-GCL-$\boldsymbol{s}_3\boldsymbol{s}_1$ &90.7&96.6&98.4&85.8\\
				CS-GCL-$\boldsymbol{s}_3\boldsymbol{s}_1\boldsymbol{s}_2$ &91.4&96.8&98.5&86.7\\
				\hline
		\end{tabular}}
	\end{center}
	\label{table5}
\end{table}

\begin{figure}[!t]
	\begin{center}
		\includegraphics[width=1.0 \linewidth]{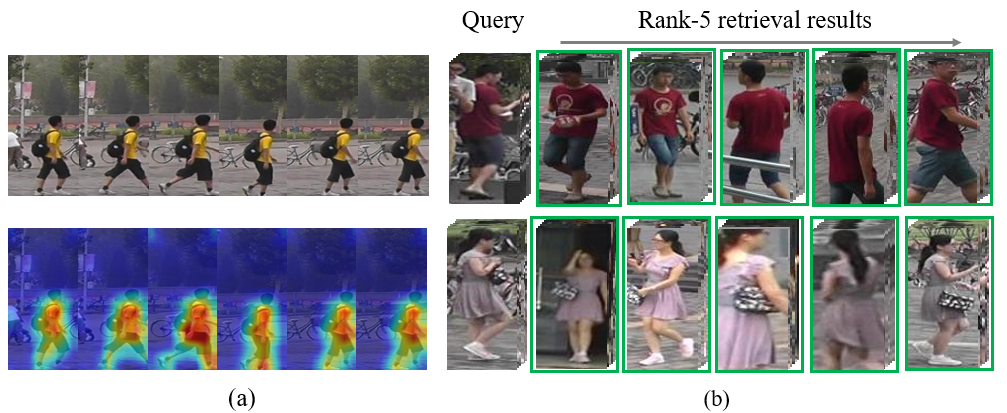}
	\end{center}
	\caption{(a) Visualization of the learned feature maps; (b) Visualization of some retrieval results by CTL.}
	\label{img8}
\end{figure}
 
\textbf{Analysis of CS-GCL.} In Table \ref{table5}, we investigate the influence of the number $M$ of CS-GCLs for transferring each granularity-level part features from one scale to another one, and analyze the performance of fusing different scales of part features. We can observe that the two-layer CS-GCLs (denotes using another CS-GCL after the first 3D-GCL) obtain performance degradation over one-layer CS-GCL. It indicates that two-layer CS-GCLs tend to fuse much redundant information, which weaken the representational capability. Moreover, CS-GCL-$\boldsymbol{s}_3\boldsymbol{s}_1\boldsymbol{s}_2$ achieves better results over CS-GCL-$\boldsymbol{s}_3\boldsymbol{s}_1$ and CS-GCL-$\boldsymbol{s}_3$ by combing more granularity-level part features. The improvement verifies CS-GCL can effectively mine the distinct patterns from each scale and enhance feature representation by fusing the complementary information among them.    

\textbf{Visualization Results.} We visualize the learned feature respond maps of one video sequence by Grad-CAM \cite{selvaraju2017grad}. In Figure \ref{img8}(a), we can observe that the feature maps from different video frames of a pedestrian have stronger response on the same discriminative regions, which verifies that CTL can extract aligned discriminative clues by modeling cross-scale spatial-temporal correlation. Figure \ref{img8}(b) shows the retrieval results of two pedestrians by CTL. We can observe that Rank-$5$ retrieval results by CTL are all matching. This indicates CTL effectively alleviates the problem of misalignment and occlusion, viewpoint variation, \textit{etc}.

\section{Conclusion}
In this work, we propose a novel Spatial-Temporal Correlation and Topology Learning framework (CTL) for video-based person re-identification to learn discriminative and robust representation. CTL utilizes a key-points estimator to extract multi-scale part features as graph nodes. A context-reinforced topology is then explored to structure multi-scale graphs by considering global contextual information and physical connection of human body. Moreover, a 3D graph convolution and a cross-scale graph convolution are designed, and performed on the multi-scale graphs. They facilitate direct cross-spacetime and cross-scale information propagation among graph nodes, and model complex relation and structural information to refine pedestrian representation. Extensive experiments on two video datasets validate the effectiveness of the proposed method.

\section*{Acknowledgment}
This work was supported by the National Key R\&D Program of China under Grand 2020AAA0105702, National Natural Science Foundation of China (NSFC) under Grants U19B2038 and U19B2023, and China Postdoctoral Science Foundation Funded Project under Grant 2020M671898.

{\small
\bibliographystyle{ieee_fullname}
\bibliography{egbib}
}

\end{document}